\documentclass{article}

\usepackage{arxiv}

\usepackage[utf8]{inputenc} 
\usepackage[T1]{fontenc}    
\usepackage{hyperref}       
\usepackage{url}            
\usepackage{booktabs}       
\usepackage{amsfonts}       
\usepackage{nicefrac}       
\usepackage{microtype}      
\usepackage{lipsum}		
\usepackage{graphicx}
\usepackage{natbib}
\usepackage{doi}
\usepackage{wrapfig}

\graphicspath{ {./images/} }

\title{Sim2Real Docs: Domain Randomization for Documents in Natural Scenes using Ray-traced Rendering}

\author{
Nikhil Maddikunta, Huijun Zhao, Sumit Keswani, Alfy Samuel, 
\\
\textbf{
Fu-Ming Guo, Nishan Srishankar, Vishwa Pardeshi, 
Austin Huang\thanks{Correspondence: austin.huang@fmr.com}
}
\\
Fidelity Investments, Artificial Intelligence Center of Excellence
}

\renewcommand{\shorttitle}{}

\hypersetup{
pdftitle={Sim2Real Docs: Domain Randomization for Documents in Natural Scenes using Ray-traced Rendering},
pdfsubject={},
pdfauthor={Nikhil Maddikunta, Huijun Zhao, Sumit Keswani, Alfy Samuel,Fu-Ming Guo, Nishan Srishankar, Vishwa Pardeshi, Austin Huang},
pdfkeywords={computer vision, documents, sim2real, data centric ai},
}

\begin{document}
\maketitle
\begin{abstract}

In the past, computer vision systems for digitized documents could rely on systematically captured, high-quality scans. Today, transactions involving digital documents are more likely to start as mobile phone photo uploads taken by non-professionals. As such, computer vision for document automation must now account for documents captured in natural scene contexts. An additional challenge is that task objectives for document processing can be highly use-case specific, which makes publicly-available datasets limited in their utility, while manual data labeling is also costly and poorly translates between use cases.

To address these issues we created Sim2Real Docs \footnote{\texttt{https://github.com/fidelity/sim2real-docs}} - a framework for synthesizing datasets and performing domain randomization of documents in natural scenes. Sim2Real Docs enables programmatic 3D rendering of documents using Blender, an open source tool for 3D modeling and ray-traced rendering. By using rendering that simulates physical interactions of light, geometry, camera, and background, we synthesize datasets of documents in a natural scene context. Each render is paired with use-case specific ground truth data specifying latent characteristics of interest, producing unlimited fit-for-task training data. The role of machine learning models is then to solve the inverse problem posed by the rendering pipeline. Such models can be further iterated upon with real-world data by either fine tuning or making adjustments to domain randomization parameters.   
\end{abstract}


\section{Introduction}

While computer vision models for extracting text from images have existed for decades (\cite{lecun1989backpropagation, tang1994document, tang1996automatic, sinha1995devanagari}), the development of new, robust neural networks for text layout, detection and recognition is still an active area of research (\cite{xu2020layoutlm, li2021trocr}), particularly in the context of text in natural scenes (\cite{bartz2018see}). 

At the same time, transactions requiring document processing in commercial applications have mostly transitioned to taking place online. Whereas previously, computer vision for document processing could rely on systematic scans, today documents are more likely to be photographed by customers themselves on a mobile phone camera under natural scene conditions. 

In light of this, computer vision objectives for document processing that were once considered solved (\cite{bartz2018see}) need to be re-examined. Classical document processing systems remain fragile - working under narrow acquisition conditions and often producing noisy data (\cite{chiron2017impact, mokhtar2018ocr}). Documents in natural scene photographs violate the fragile operating parameters of these traditional document processing pipelines. 

Recent advances in neural networks for text recognition in natural scenes (\cite{mishraICDAR19, hu2020iterative, lyu2018mask}) as well as layout analysis and recognition using transformer architectures (\cite{xu2020layoutlm, li2021trocr}) suggest a promising direction which robustly unifies text and vision in natural scenes.

However, a major impediment to taking full advantage of neural networks in this context is the availability of training data. An additional challenge is that document processing use cases can have highly specialized task objectives, rendering shared public datasets of limited use for creating neural networks with a specific end-to-end task objective. Recent tools for synthetic document data such as \cite{gupte2021lights} are limited to applying compositions of basic image augmentations - blur, bleed-through, morphological operations - which are insufficient to represent the physical interactions (e.g. light and shadow, geometry, camera) underlying image variation in a natural scene contexts.  

We bridge this data gap by creating Sim2Real Docs (\texttt{https://github.com/fidelity/sim2real-docs}) - a Python framework which integrates with the open source Blender project. Blender is a 3D tool for modeling and rendering that is often used in movies, animations, and game development (\cite{blender}). 

\section{Synthesis Pipeline: Target Specification, Content and Scene Randomization, Rendering}

Building datasets and models using Sim2Real Docs takes an "inverse problem" view (\cite{de2005learning}) of document processing. The objective of the inverse problem is to infer a latent state corresponding to a computable representation (classification, segmentation, orientation, detection, recognition, content extraction etc.) from raw images of documents in natural scenes. 

\begin{figure}[t]
\label{fig:render}
\caption{Render (left), depth map (middle), and in-camera document locations for fields of interest (right).}
\includegraphics[width=3cm]{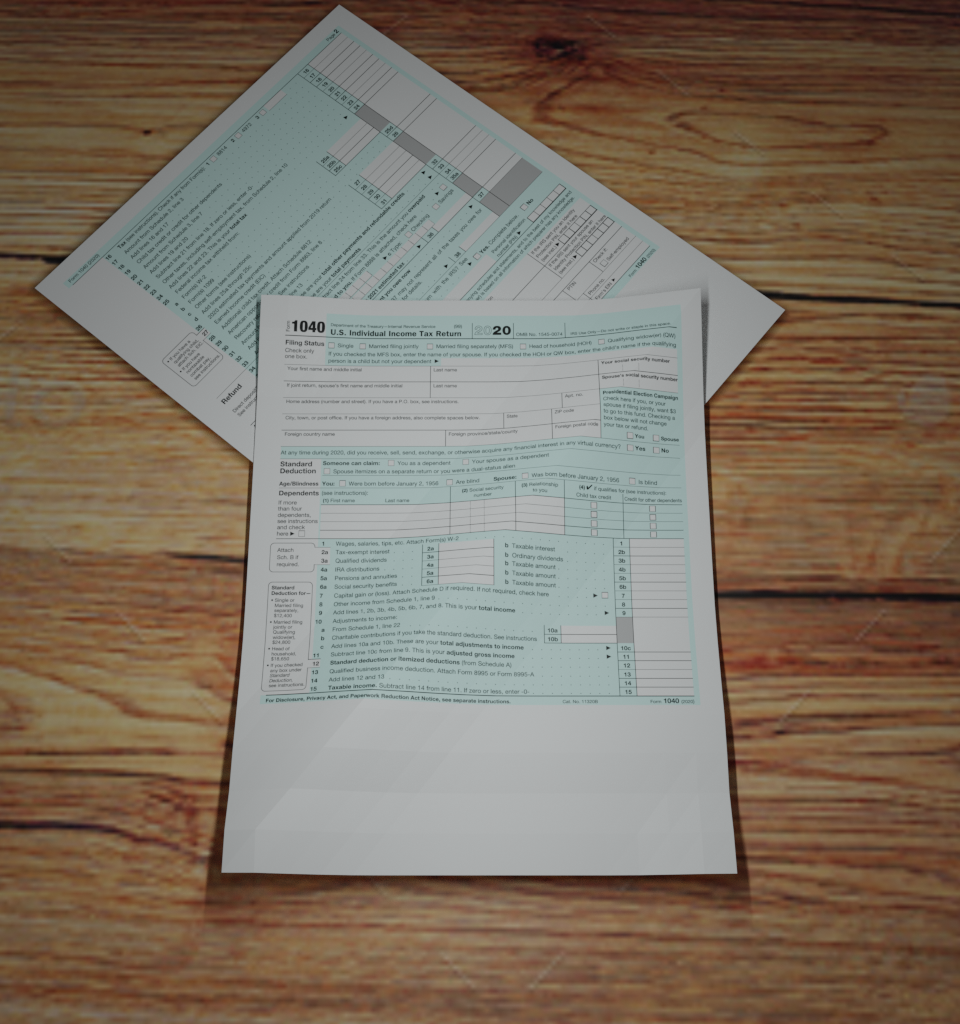}
\includegraphics[width=3cm]{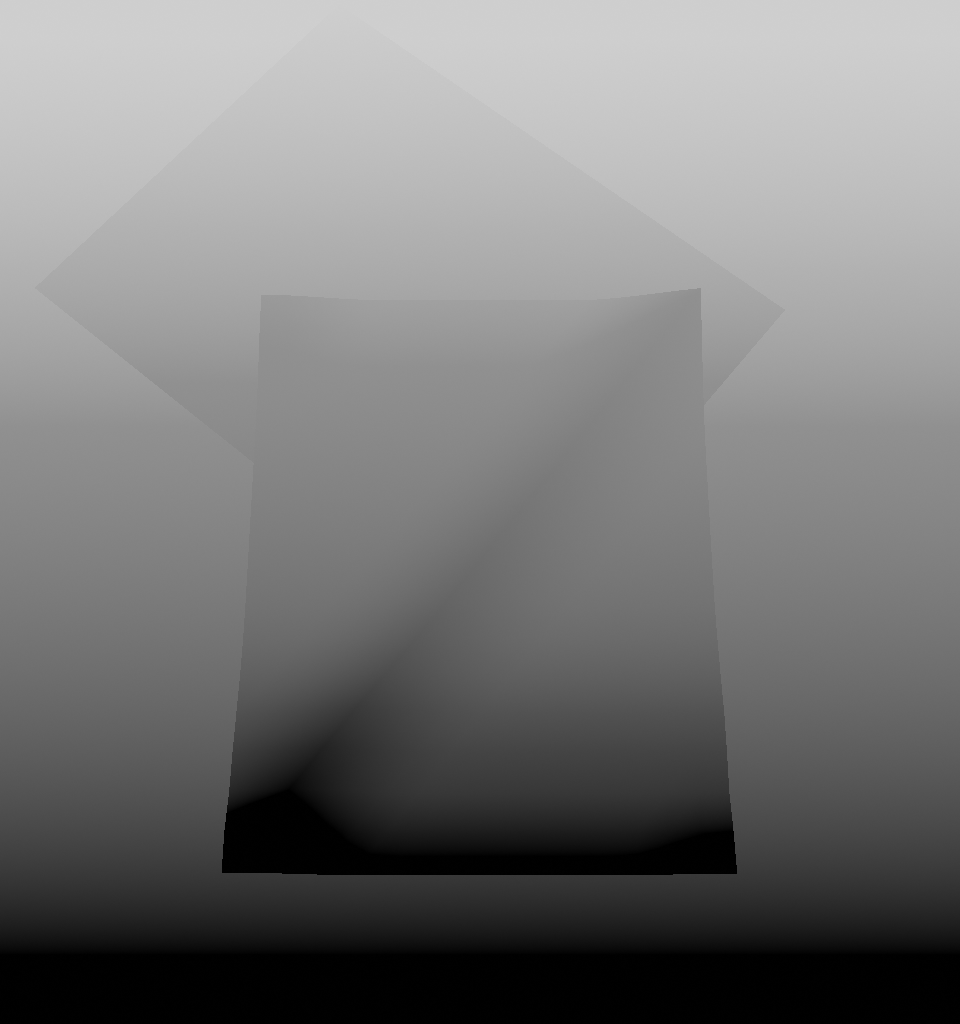}
\includegraphics[width=3cm]{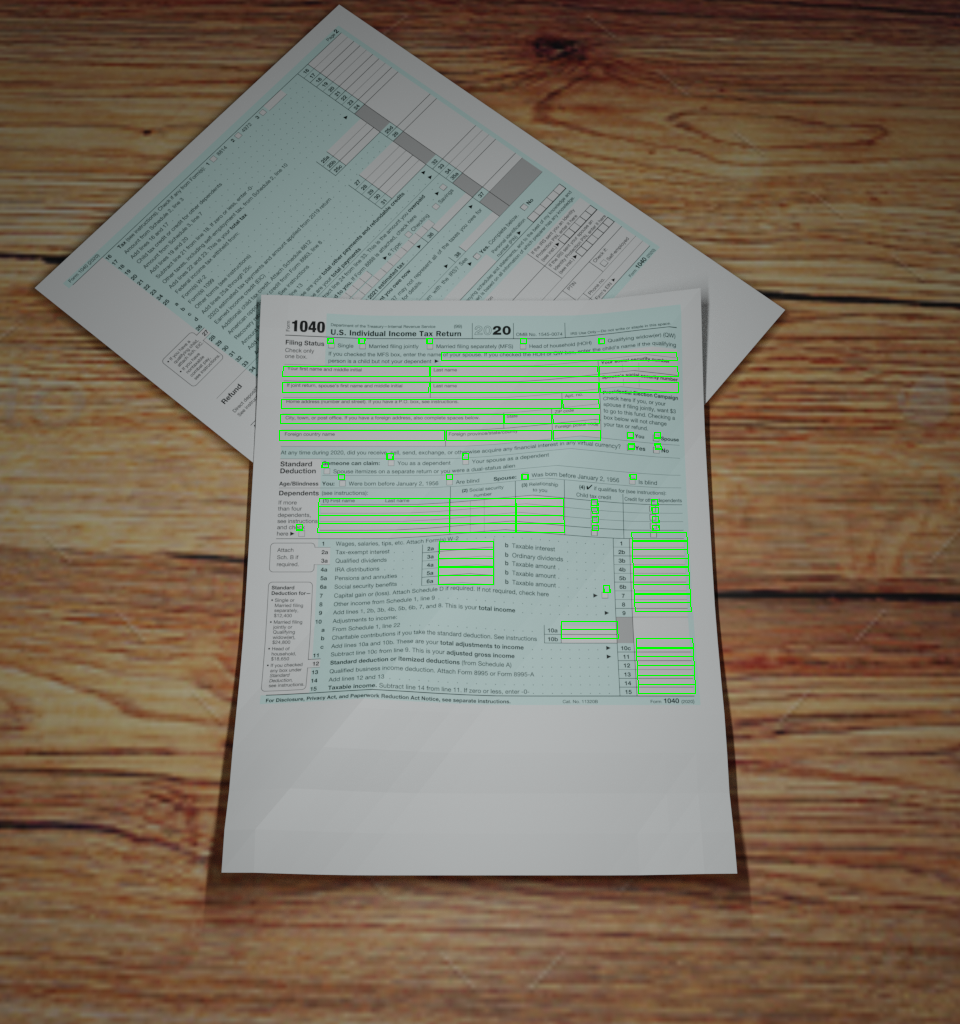}
\centering
\end{figure}

Simulation via rendering implements the "forward" computation of the inverse problem - synthesis starts from the latent state value to produce a rendered image, with domain randomized scene parameters to generate variation reflecting class invariants (for example, document classification should be invariant to lighting or camera angles). The role of the neural network is to learn a program which solves the end-to-end inverse problem posed by the simulated scene and rendering process.   

By using a data synthesis approach with a photorealistic ray-traced rendering engine embedded in the domain randomization process, the ground truth is always available as part of the generated scene specification. Furthermore, the physics embedded in the rendering engine enables a layer of learned invariants that would be difficult to specify directly using standard data augmentation techniques.   

\begin{figure}[t]
\label{fig:image_grid}
\caption{Sample images from a synthesized classification dataset.}
\includegraphics[width=16cm]{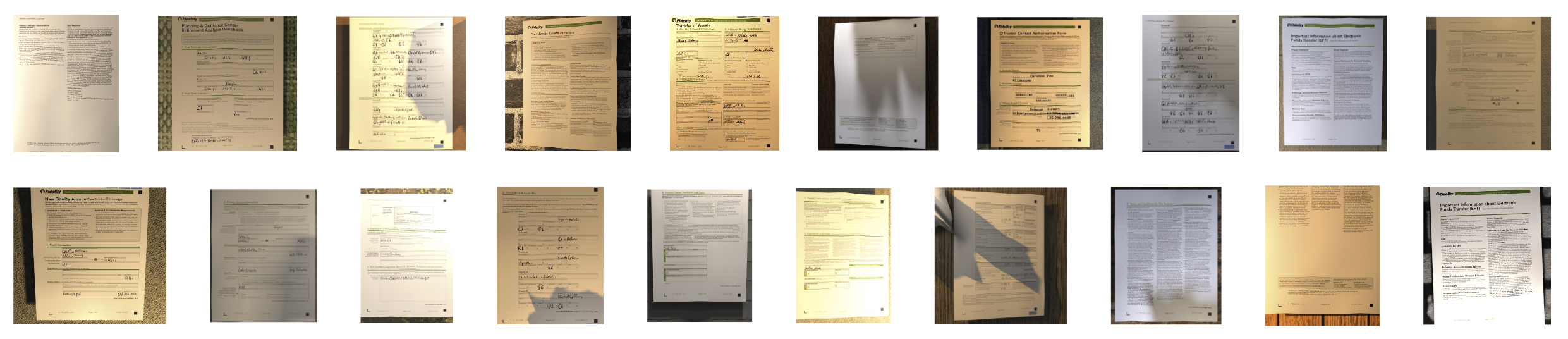}
\centering
\end{figure}

With this in mind, specifying a model task with Sim2Real Docs consists of the following steps:

\begin{enumerate}
    \item \textbf{Define and randomly sample base document images (e.g. clean pdfs) for the model supervision task}. Some examples of task objectives are described in the following section ("Example Use Cases"). 
    \item \textbf{Content randomization (optional)}. To ensure vision models learn to be content-invariant, documents can be randomly populated with fake content (e.g. \cite{Faraglia_Faker}), either using typed text or handwritten content from a Generative Adversarial Network (GAN) such as \cite{kang2020ganwriting}.
    \item \textbf{Style noise (optional)}. When it's necessary to capture texture quality of documents such as faxes, photocopies, scanning, etc., style transfer can be randomly applied to a subset of documents \cite{Gatys2015c}. Alternatively classical noise models such as Gaussian noise and morphological operations as in (\cite{gupte2021lights}) can also be applied. 
    \item \textbf{Scene randomization}. Domain randomization is performed with respect to the scene contents, which include a document image with optional modifications from steps (2) and (3) applied as a texture to a 3D document object. Domain randomization of the scene teaches invariants to the model such as camera position and orientation relative to documents, camera lens properties such as focus and depth of field, scene background (e.g.  table surface, other pages, incidental objects), document geometry (folds, bends, creases in the paper), lighting (sources, intensity, shadows), occlusions and cropping.
    \item \textbf{Rendering engine invocation}. For each generated scene, Sim2Real Docs invokes the path-traced rendering process of the Cycles engine to produce a final image (\ref{fig:image_grid}). 
\end{enumerate}

\section{Example Use Cases}
\label{examples}

Here we describe several example use cases and illustrate some approaches to address them.

\textbf{Camera Rotation}. A common issue with document images is for the orientation to be flipped, sideways, or rotated at an angle. The Nvidia OCRRot library (\cite{nvidia_ocrrot}) implements a neural network to rotate pages to their correct orientation, but only for multiples of 90 degrees. 

Using Sim2RealDocs, ground truth data is produced by randomizing camera rotation, while domain randomization is applied to other parameters such as  camera translation, lighting, shadows, and other variables. In contrast to using rotation data augmentation (such as in torchvision), boundary artifacts are absent because an actual simulated camera is undergoing rotation. For estimating a rotation angles, boundary artifacts would constitute a form of data leakage. Using the synthesized data, any standard vision backbone (\cite{he2016deep}) can be used to fine tune output with a periodic loss: 

\newcommand{\taninv}{\tan^{-1}}
\begin{equation}
    \mathcal{L} = [sin(\hat{\theta}) - sin(\theta)]^2 + [cos(\hat{\theta}) - cos(\theta)]^2
\end{equation}

$\theta$ is the ground truth camera rotation and the model outputs the $sin(\hat{\theta})$ and $cos(\hat{\theta})$ predictions. 

\textbf{Document Classification}. Documents often need to be sorted into one of a set of known document types, with allowances for an "unknown/other" category. This determination must be invariant to variations in content, orientation, lighting, placement, and background (Figure \ref{fig:image_grid}). This can be approached as a standard classification problem with some caveats. If document classes are similar in appearance, the image classifier may need to incorporate information that captures fine detail. If documents consist of more than one page, each document-page pair is its own class. One advantage of performing document classification upstream of OCR is that inference latency can be realtime and classification is not subject to OCR errors.

\textbf{Logo Detection}. It's common to identify brands in documents by detecting their logos/imagery. For base images, one can either obtain documents with example logos or insert logos as a form of content augmentation (step 2 of "Synthesis Pipeline"). Ground truth data is produced by localizing the logo position in the document and performing a projective transformation to obtain ground truth bounding box in the camera space projection (Figure \ref{fig:render}, right, illustrates the use of projective transformations to convert content bounding boxes to in-camera coordinates). Rendered images are then used to train a standard detection model.

\textbf{OCR Readability}. Image quality issues such as camera focus, grain/noise, low lighting conditions, fonts, poor camera angles, can cause downstream OCR systems to fail. Being able to diagnose potential issues in realtime prior to batch OCR processes allows alerting users that an image may need to be re-taken. We can use self-supervision to derive such models by defining a loss that compares the difference between OCR of the base document against OCR of the rendered image for a distribution of documents.

\textbf{Segmentation}. It's common to use simple heuristics such as thresholding for segmenting documents in the foreground from their background. However, such heuristics can fail if there are textures in the background or other pages beneath the foreground document. By generating a ground truth per-pixel segmentation, segmentation models can be taught to be invariant to background contexts that would otherwise cause heuristic methods to fail.

\textbf{View Synthesis / Document Coordinate Transforms}. There is a frequent need to transform from the camera coordinate space to the document coordinate space in order to contextualize and extract content. For example, content of interest (such as an account ID or signature) may occur at a known fixed location on the page. This transformation of a document image from the camera coordinate system to a document-space projection is a simplified instance of view synthesis. For this, the camera position and orientation relative to the document within the 3D scene provides a ground truth projective transformation (Figure \ref{fig:render} middle) that can be used for model training. A more challenging variant of this task is to apply distortions such as folds, creases, and bends to the document geometry as part of the domain randomization.

\textbf{Detection and Recognition}. Text detection and recognition models can be trained using rendered images since the base image text and any augmented document content are known and provide ground truth data. Projective transformations can be used to transform text locations within a document to camera-space bounding boxes (Figure \ref{fig:render} right).

\begin{wrapfigure}{r}{0.5\textwidth}
\label{fig:config}
\caption{Specification file excerpt defining parameter ranges for domain randomization.}
\centering
\frame{\includegraphics[width=7 cm]{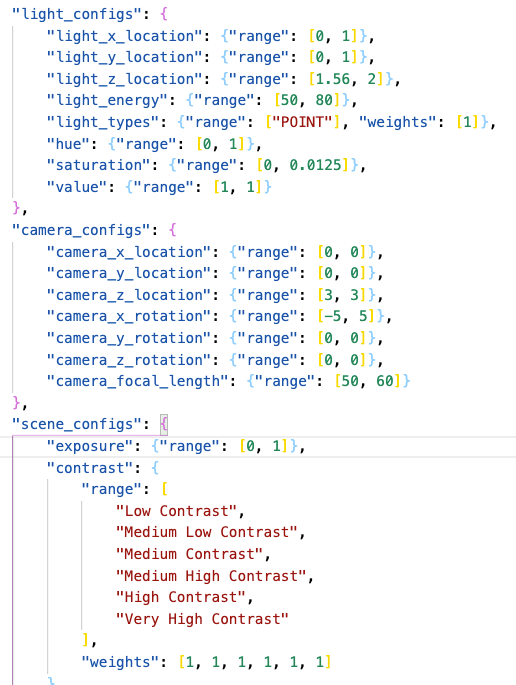}}
\end{wrapfigure}

\section{Usage}

From the user specification for a dataset (Figure \ref{fig:config}) , Sim2Real Docs randomizes a 3D scene for each image in the input location, creating random scene variations according to ranges (for continuous parameters) and user-weighted distributions (for categorical parameters). Running the Sim2Real Docs process creates a scene per base image, renders it, and produces accompanying metadata reflecting the scene conditions. Sim2Real Docs can also return a tuple of Blender object models for each render describing the camera, lighting, and scene for further scripting or customization.

\section{Conclusion}

Sim2Real Docs enables synthetic data and domain randomization for documents in natural scenes using ray-traced rendering. Synthetic data provides a direct approach to data-centric model development - a task is defined by creating data where the model target variable is a parameter of the simulation's specification. With photorealism becoming increasingly accessible using tools such as Blender (\cite{blender}), Unreal Engine (\cite{unrealengine}), as well as machine learning methods such as GANs (\cite{richter2021enhancing}), we expect simulation approaches for data-centric AI to become increasingly common. 

In the past, text in natural scenes was considered a distinct domain from text in documents. Beyond the immediate applications of Sim2Real Docs, our broader goal is to eliminate this distinction and unify content extraction and comprehension for documents and natural scenes with multimodal models that are more general and robust.

\section{Acknowledgements}

The authors would like to thank Qian Yang for comments and edits to this manuscript.

\bibliographystyle{unsrtnat}
\bibliography{references}  

\begin{thebibliography}{22}
\providecommand{\natexlab}[1]{#1}
\providecommand{\url}[1]{\texttt{#1}}
\expandafter\ifx\csname urlstyle\endcsname\relax
  \providecommand{\doi}[1]{doi: #1}\else
  \providecommand{\doi}{doi: \begingroup \urlstyle{rm}\Url}\fi

\bibitem[LeCun et~al.(1989)LeCun, Boser, Denker, Henderson, Howard, Hubbard,
  and Jackel]{lecun1989backpropagation}
Yann LeCun, Bernhard Boser, John~S Denker, Donnie Henderson, Richard~E Howard,
  Wayne Hubbard, and Lawrence~D Jackel.
\newblock Backpropagation applied to handwritten zip code recognition.
\newblock \emph{Neural computation}, 1\penalty0 (4):\penalty0 541--551, 1989.

\bibitem[Tang et~al.(1994)Tang, De~Yan, and Suen]{tang1994document}
Yuan~Yan Tang, Chang De~Yan, and Ching~Y. Suen.
\newblock Document processing for automatic knowledge acquisition.
\newblock \emph{IEEE transactions on Knowledge and Data Engineering},
  6\penalty0 (1):\penalty0 3--21, 1994.

\bibitem[Tang et~al.(1996)Tang, Lee, and Suen]{tang1996automatic}
Yuan~Y Tang, Seong-Whan Lee, and Ching~Y Suen.
\newblock Automatic document processing: a survey.
\newblock \emph{Pattern recognition}, 29\penalty0 (12):\penalty0 1931--1952,
  1996.

\bibitem[Sinha and Bansal(1995)]{sinha1995devanagari}
RMK Sinha and Veena Bansal.
\newblock On devanagari document processing.
\newblock In \emph{1995 IEEE International Conference on Systems, Man and
  Cybernetics. Intelligent Systems for the 21st Century}, volume~2, pages
  1621--1626. IEEE, 1995.

\bibitem[Xu et~al.(2020)Xu, Li, Cui, Huang, Wei, and Zhou]{xu2020layoutlm}
Yiheng Xu, Minghao Li, Lei Cui, Shaohan Huang, Furu Wei, and Ming Zhou.
\newblock Layoutlm: Pre-training of text and layout for document image
  understanding.
\newblock In \emph{Proceedings of the 26th ACM SIGKDD International Conference
  on Knowledge Discovery \& Data Mining}, pages 1192--1200, 2020.

\bibitem[Li et~al.(2021)Li, Lv, Cui, Lu, Florencio, Zhang, Li, and
  Wei]{li2021trocr}
Minghao Li, Tengchao Lv, Lei Cui, Yijuan Lu, Dinei Florencio, Cha Zhang,
  Zhoujun Li, and Furu Wei.
\newblock Trocr: Transformer-based optical character recognition with
  pre-trained models.
\newblock \emph{arXiv preprint arXiv:2109.10282}, 2021.

\bibitem[Bartz et~al.(2018)Bartz, Yang, and Meinel]{bartz2018see}
Christian Bartz, Haojin Yang, and Christoph Meinel.
\newblock See: towards semi-supervised end-to-end scene text recognition.
\newblock In \emph{Thirty-second aaai conference on artificial intelligence},
  2018.

\bibitem[Chiron et~al.(2017)Chiron, Doucet, Coustaty, Visani, and
  Moreux]{chiron2017impact}
Guillaume Chiron, Antoine Doucet, Micka{\"e}l Coustaty, Muriel Visani, and
  Jean-Philippe Moreux.
\newblock Impact of ocr errors on the use of digital libraries: towards a
  better access to information.
\newblock In \emph{2017 ACM/IEEE Joint Conference on Digital Libraries (JCDL)},
  pages 1--4. IEEE, 2017.

\bibitem[Mokhtar et~al.(2018)Mokhtar, Bukhari, and Dengel]{mokhtar2018ocr}
Kareem Mokhtar, Syed~Saqib Bukhari, and Andreas Dengel.
\newblock Ocr error correction: State-of-the-art vs an nmt-based approach.
\newblock In \emph{2018 13th IAPR International Workshop on Document Analysis
  Systems (DAS)}, pages 429--434. IEEE, 2018.

\bibitem[Mishra et~al.(2019)Mishra, Shekhar, Singh, and
  Chakraborty]{mishraICDAR19}
Anand Mishra, Shashank Shekhar, Ajeet~Kumar Singh, and Anirban Chakraborty.
\newblock Ocr-vqa: Visual question answering by reading text in images.
\newblock In \emph{ICDAR}, 2019.

\bibitem[Hu et~al.(2020)Hu, Singh, Darrell, and Rohrbach]{hu2020iterative}
Ronghang Hu, Amanpreet Singh, Trevor Darrell, and Marcus Rohrbach.
\newblock Iterative answer prediction with pointer-augmented multimodal
  transformers for textvqa.
\newblock In \emph{Proceedings of the IEEE/CVF Conference on Computer Vision
  and Pattern Recognition}, pages 9992--10002, 2020.

\bibitem[Lyu et~al.(2018)Lyu, Liao, Yao, Wu, and Bai]{lyu2018mask}
Pengyuan Lyu, Minghui Liao, Cong Yao, Wenhao Wu, and Xiang Bai.
\newblock Mask textspotter: An end-to-end trainable neural network for spotting
  text with arbitrary shapes.
\newblock In \emph{Proceedings of the European Conference on Computer Vision
  (ECCV)}, pages 67--83, 2018.

\bibitem[Gupte et~al.(2021)Gupte, Romanov, Mantravadi, Banda, Liu, Khan,
  Meenal, Han, and Srinivasan]{gupte2021lights}
Amit Gupte, Alexey Romanov, Sahitya Mantravadi, Dalitso Banda, Jianjie Liu,
  Raza Khan, Lakshmanan~Ramu Meenal, Benjamin Han, and Soundar Srinivasan.
\newblock Lights, camera, action! {A} framework to improve {NLP} accuracy over
  {OCR} documents.
\newblock \emph{CoRR}, abs/2108.02899, 2021.
\newblock URL \url{https://arxiv.org/abs/2108.02899}.

\bibitem[Community(2021)]{blender}
Blender~Online Community.
\newblock \emph{Blender - a 3D modelling and rendering package}.
\newblock Blender Foundation, Stichting Blender Foundation, Amsterdam, 2021.
\newblock URL \url{http://www.blender.org}.

\bibitem[De~Vito et~al.(2005)De~Vito, Rosasco, Caponnetto, De~Giovannini,
  Odone, and Bartlett]{de2005learning}
Ernesto De~Vito, Lorenzo Rosasco, Andrea Caponnetto, Umberto De~Giovannini,
  Francesca Odone, and Peter Bartlett.
\newblock Learning from examples as an inverse problem.
\newblock \emph{Journal of Machine Learning Research}, 6\penalty0 (5), 2005.

\bibitem[Faraglia and {Other Contributors}(2021)]{Faraglia_Faker}
Daniele Faraglia and {Other Contributors}.
\newblock {Faker}, 2021.
\newblock URL \url{https://github.com/joke2k/faker}.

\bibitem[Kang et~al.(2020)Kang, Riba, Wang, Rusi{\~n}ol, Forn{\'e}s, and
  Villegas]{kang2020ganwriting}
Lei Kang, Pau Riba, Yaxing Wang, Mar{\c{c}}al Rusi{\~n}ol, Alicia Forn{\'e}s,
  and Mauricio Villegas.
\newblock Ganwriting: Content-conditioned generation of styled handwritten word
  images.
\newblock In \emph{European Conference on Computer Vision}, pages 273--289.
  Springer, 2020.

\bibitem[Gatys et~al.(2015)Gatys, Ecker, and Bethge]{Gatys2015c}
Leon~A Gatys, Alexander~S Ecker, and Matthias Bethge.
\newblock A neural algorithm of artistic style.
\newblock \emph{arXiv preprint arXiv:1508.06576}, 2015.

\bibitem[Nvidia(2018)]{nvidia_ocrrot}
Nvidia.
\newblock ocrrot, 2018.
\newblock URL \url{https://github.com/NVlabs/ocropus3-ocrorot}.

\bibitem[He et~al.(2016)He, Zhang, Ren, and Sun]{he2016deep}
Kaiming He, Xiangyu Zhang, Shaoqing Ren, and Jian Sun.
\newblock Deep residual learning for image recognition.
\newblock In \emph{Proceedings of the IEEE conference on computer vision and
  pattern recognition}, pages 770--778, 2016.

\bibitem[{Epic Games}(2021)]{unrealengine}
{Epic Games}.
\newblock Unreal engine, 2021.
\newblock URL \url{https://www.unrealengine.com}.

\bibitem[Richter et~al.(2021)Richter, AlHaija, and
  Koltun]{richter2021enhancing}
Stephan~R Richter, Hassan~Abu AlHaija, and Vladlen Koltun.
\newblock Enhancing photorealism enhancement.
\newblock \emph{arXiv preprint arXiv:2105.04619}, 2021.

\end{thebibliography}


\end{document}